\documentclass{article}

\usepackage{PRIMEarxiv}

\usepackage[utf8]{inputenc} 
\usepackage[T1]{fontenc}    
\usepackage{hyperref}       
\usepackage{xcolor}
\hypersetup{
    colorlinks,
    linkcolor={red!50!black},
    citecolor={blue!50!black},
    urlcolor={blue!80!black}
}
\usepackage{url}            
\usepackage{booktabs}       
\usepackage{amsfonts}       
\usepackage{nicefrac}       
\usepackage{microtype}      
\usepackage{lipsum}
\usepackage{fancyhdr}       
\usepackage{graphicx}       

\usepackage{amsmath}
\usepackage{amssymb}
\usepackage{mathtools}
\usepackage{amsthm}

\usepackage{bm}
\usepackage{subfigure}
\usepackage{natbib}
\usepackage{doi}
\usepackage[capitalize,noabbrev]{cleveref}

\theoremstyle{plain}
\newtheorem{theorem}{Theorem}[section]

\theoremstyle{definition}

\theoremstyle{remark}

\usepackage{multicol}

\title{An Interpretable Determinantal Choice Model for Subset Selection
}

\author{Sander Aarts \thanks{School of Operations Research and Information Engineering, Cornell University. Correspondence: \texttt{sea78@cornell.edu}.} \and David B. Shmoys \footnotemark[1] \and Alex Coy \thanks{School of Electrical and Computer Engineering, Cornell University}}

\newcommand{\mcU}{\mathcal{U}}

\newcommand{\bfL}{\mathbf{L}}
\newcommand{\bfS}{\mathbf{S}}
\newcommand{\bfK}{\mathbf{K}}
\newcommand{\bfX}{\mathbf{X}}
\newcommand{\bfx}{\mathbf{x}}

\newcommand{\bfy}{\mathbf{y}}

\newcommand{\bfz}{\mathbf{z}}
\newcommand{\bfI}{\mathbf{I}}
\newcommand{\bfJ}{\mathbf{J}}

\DeclareMathOperator{\mcc}{MCC}

\usepackage{url}

\begin{document}
\maketitle

\begin{abstract}
    Understanding how subsets of items are chosen from offered sets is critical to assortment planning, wireless network planning, and many other applications. There are two seemingly unrelated subset choice models that capture dependencies between items: intuitive and interpretable random utility models; and tractable determinantal point processes (DPPs). This paper connects the two. First, all DPPs are shown to be random utility models. Next, a determinantal choice model that enjoys the best of both worlds is specified; the model is shown to subsume logistic regression when dependence is minimal, and MNL when dependence is maximally negative. This makes the model interpretable, while retaining the tractability of DPPs. A simulation study verifies that the model can learn a continuum of negative dependencies from data, and an applied study using original experimental data produces novel insights on wireless interference in LoRa networks.
\end{abstract}



\section{Introduction}
\label{Introduction}

Assortment planning is one of many applications in which it is critical to understand how subsets of items are selected out of a potentially varying set of offered items. In assortment planning a retailer offers a personalized assortment for each customer to browse and make purchases from. Effective planning requires understanding how likely a customer is to purchase a subset of an assortment \emph{as a function of the assortment}. Another application is predicting returns of online orders; given an order, what subset of items will likely be returned? Finally, in wireless network planning -- given a set of wireless transmissions within a time-interval -- it is critical to predict what subset of transmissions are successfully received and what subset is lost. Understanding the underlying mechanism is critical for network design. In fact, this last setting originally motivated our work, and has provided the test bed for many of our experiments. Common to these applications is the need for a \textit{subset choice model}. In addition, the model should be interpretable in order for planners understand the underlying mechanisms of subset selection. Naturally, it should also be learnable from data.

Traditional \textit{discrete choice models} assume that at most a single item is selected. A celebrated example is the multinomial logit model (MNL) \citep{McFadden73}. The model admits efficient learning algorithms for its parameters, and is both intuitive and interpretable. This is in part due to the model being founded on random utility maximization (RUM); each item is equipped with a utility function that maps an item's features to a utility value that quantifies its relative desirability. The most desirable alternative is chosen. The parameters of the utility function are interpretable as marginal utilities. Furthermore, the model is parsimonious; given a set of parameters, the model can generate predictions over assortments of varying sizes \citep{GallegoTopaloglu2019}. The restriction to single-item selections, however, limits the applicability of MNL to subset choice. Clearly, subset choice provides a more general model.

Generalizing discrete choice to subset choice is challenging due to item \emph{interactions} within subsets. Items are said to interact whenever the utility of a subset differs from the sum of utilities of the items it contains. For example, the set $\{\texttt{toothbrush}, \texttt{toothpaste}\}$ might generate more utility than the sum of utilities of singletons $\{\texttt{toothbrush}\}$ and $\{\texttt{toothpaste}\}$. This is a positive interaction. Meanwhile, unrelated items such as $\{\texttt{toothpaste}, \texttt{pencil}\}$ may not interact. Finally, a set such as $\{\texttt{electric toothbrush}, \texttt{toothbrush}\}$ likely produces less utility than the sum of its parts; this is a negative interaction. These are particularly common when items are substitutes. Note that item interactions are not limited to subsets of size two; sets of arbitrary size can interact. Capturing arbitrary interactions in a model, however, is not generally tractable. 

\emph{Subset choice models} strike a balance between tractability and the generality of the interactions they can capture. An early example is the additive model \citep{Fishburn1992}. This model considers no interactions, and models the utility of a set as the sum of utilities of the individual items it contains. Most subset choice models are ``additive with corrections'' -- the utility of a subset is additive in individual utilities, with some subsets receiving some form of ``correction''. Examples include giving each set pairwise corrections \citep{Fishburn1996}, or giving a limited number of sets arbitrary corrections \citep{BensonSubset2018}. The former model is not used in learning contexts; the latter is. Moreover, some subset choice models  are not founded on RUM at all.

Determinantal point processes (DPPs) are a family of subset selection models used heavily in machine learning contexts. While not founded on random utility maximization, DPPs are remarkably tractable for learning and inference tasks in which items exhibit repulsion; they were originally used to model repulsion between fermions \citep{macchi1975}. DPPs admit efficient learning algorithms, and support inference tasks such as sampling and conditioning (see, e.g., \citet{KuleszaTaskarBook}). DPPs are successfully employed in learning tasks such as document summarization \citep{KuleszaTaskarLearningDPPs}, basket completion \citep{GartellBayesLowRank}, and wireless interference \citep{SahaDhillon2019}. However, unlike RUM-based models, DPPs are  less interpretable, and do not enjoy the same intuitive appeal as their utility-based counterparts.

This paper shows that determinantal processes are in fact subset choice models founded on random utility maximization, and specifies a sub-family of DPPs that enjoys strong connections to well-known discrete choice models. These connections make DPPs more intuitive, interpretable, and accessible to practitioners with backgrounds in discrete choice modeling. Our contributions are the following: (1) We prove that \emph{any} determinantal processes is equivalent to an MNL model over subsets, with a particular utility function; (2) We prove that this utility function is ``additive with corrections'' where each set receives a corrective penalty for containing similar items; (3) We specify a \textit{determinantal choice model} and connect it to discrete choice, in particular; (4) We show that our model subsumes logistic regression when there are no item interactions, and; (5) that the model subsumes MNL when the interactions are maximally negative.
Moreover, we showcase the determinantal choice model in practice: (6) We use synthetic data to verify that the determinantal choice model, when learned from data, subsumes logistic regression and MNL as expected; (7) Finally, we use original LoRaWAN wireless data to learn a wireless interference model. Our results show that the model both captures the negative dependence in the data, and learns interpretable parameters. Indeed, we use these parameters to draw new insights on LoRaWAN transmission loss due to interference.


\section{Setting and Related Work}
\label{sec:background}
This section introduces the setting for subset choice under variable assortments, and reviews discrete choice, random utility maximization, subset choice models, and DPPs.

\textbf{Problem setting.} This work adopts a feature-based parametrization and learning approach to subset choice. Consider an arbitrary universe of items $\mcU$ in which each item $i \in \mcU$ has an associated feature vector $\bfx_i \in \mathbb{R}^d$. An \emph{assortment} is a finite set of items, $A \subseteq \mcU$. The feature vectors of assortment, $A$, are collected in the $|A| \times d$ matrix $\bfX_A$. A \emph{subset choice model} is a function that takes any assortment $A$ with features $\bfX_A$ and returns a probability distribution over the collection of subsets $2^A$.  We consider a supervised learning setting, in which access to a sequence, $k = 1, \dots, K$, of assortments $A_k$, choices $C_k \subseteq A_k$, and features $\bfX_k = \bfX_{A_k}$, is assumed. Note that the assortments can differ in size, and may contain entirely different items from one another\footnote{The features can also encode assortment-level features. The values of such features are the same across all items in the assortment. This can capture, e.g., features about the choice-maker, or the time and place at which the assortment was offered.}.  Prediction involves inferring a likely choice set $C_k \subseteq A_k$, given the assortment $A_k$, and collection of associated item features $\bfX_k$.

\textbf{Discrete choice and MNL.} Discrete choice models allow at most one item to be selected from an assortment. Many of the most common discrete choice models are founded on random utility maximization (see, e.g., \citet{Train2009}). In this setting every item $i$ in an assortment $A$ is equipped with a real-valued utility $u(i)$, and the utility-maximizing choice is selected. Randomness in the utilities generates randomness in the selection. The MNL model is a prominent example \citep{McFadden73}. In an MNL model, the utility of item $i$ has a linear feature-based parametric form
\begin{equation*}
    u(i) = \bm\beta^T \bfx_i  + \epsilon_i, \quad \forall i \in \mcU,
\end{equation*}
where $\bm\beta$ is a coefficient vector of the same dimension as $\bfx_i$, and the $\epsilon_i$ are iid standard Gumbel random variables. Given assortment $A$, the probability of selecting item $i \in A$ is equivalent to the probability that $u(i) > u(j)$ for all $j \in A: j \neq i$, which under the Gumbel errors is proportional to the exponentiated non-random utility components.
\begin{equation}
    \label{eq:mnl-likelihood}
    \Pr[\textrm{choose item } i \in A] = 
    \frac{\exp\{\bm{\beta}^T\bfx_i\}}{1 + \sum_{j \in A}\exp\{\bm{\beta}^T\bfx_j\}}
\end{equation}
The ``no item'' option is treated as an item with expected utility $0$. Consequently the probability of opting out without an item is  $1 / (1 + \sum_{j \in A}\exp\{\bm\beta^T\bfx_j \})$. An important special case of the MNL model is when the assortment is a single item $\{i\}$. This reduces the alternatives to ``take it'' or ``leave it''. The probability of taking the item $i$ is then
\begin{equation}
    \label{eq:logit-likelihood}
    \Pr[\textrm{choose item } i \in A] = 
    \frac{\exp\{\bm{\beta}^T\bfx_i\}}{1 + \exp\{\bm{\beta}^T\bfx_i\}}.
\end{equation}
This model is also known as the logit model, or \emph{logistic regression}, in learning contexts.


\textbf{Subset choice models}. Subset choice models generalize discrete choice models by equipping each \textit{subset} of items with a random utility. The utility is composed of a non-random and random component, $U(C) = V(C) + \epsilon(C)$. The subset of maximum utility is selected. Even modestly sized assortments have prohibitively many subsets; most models trade off generality with tractability by modeling the non-random utility components as ``additive with corrections'',
\begin{equation}
    \label{eq:additive_correction}
    V(C) = \sum_{i \in C}v(i) + W(C), \quad \forall C \subseteq A, 
\end{equation}
where $W(C)$ is the correction to subset $C$. A well-know example is the pairwise interaction model with correction $W(C) = \sum_{i, j \in C: i \neq j}\gamma(i, j)$ \citep{Fishburn1996}. A more recent example is a learnable model that allows arbitrary corrections $W(C)$, however restricted to fixed number of subsets \citep{BensonSubset2018}. The latter model is best suited to settings in which a relatively small number of assortments are observed repeatedly.

\textbf{Determinantal point processes}. A determinantal point process (DPP) is a probability distribution over subsets that favors diverse subsets. A DPP over a fixed assortment $A$ is characterized by its \textit{kernel matrix} $\bfL$ \cite{borodin}. This is an $|A| \times |A|$ symmetric p.s.d. matrix\footnote{The abbreviation p.s.d. stands positive semi-definite. An $n \times n$ matrix $\bfK$ is p.s.d. if for all $\bfz \in \mathbb{R}^n$, it holds that $\bfz^T \bfK \bfz \geq 0$.}. The likelihood of observing subset $C \subseteq A$ is proportional to the log-determinant of the submatrix $\bfL_C$ of  the kernel $\bfL$ corresponding to the rows and columns in $C$:
\begin{equation}
\label{eq:likelihood_kernel}
    \Pr[\text{Choose } C] = \frac{\det(\bfL_C)}{\det(\bfI + \bfL)}.
\end{equation}
The determinant of $\bfL_\emptyset$ is taken to be $1$. The denominator is a closed-form expression for the normalizing constant, satisfying $\sum_{B \subseteq A}\det(\bfL_B) = \det(\bfI  + \bfL)$, where throughout $\bfI$ denotes an identity matrix of the same size as $\bfL$. DPPs support methods for supervised learning via both Bayesian inference and MLE \citep{KuleszaTaskarLearningDPPs, Affandi2014Learning}. Moreover, there are efficient sampling algorithms for DPPs \citep{HoughDPPSampling}. We refer to \citet{KuleszaTaskarBook} for more details on, and properties of DPPs.

The DPP likelihood favors diversity in the selected subset. This is demonstrated using a two-item example \citep{KuleszaTaskarBook}. The likelihood of selecting singleton $\{i\}$ is proportional to the $i$th diagonal entry of the kernel matrix.
\begin{equation*}
    \Pr[\text{Choose } \{i\}] \propto \det(\bfL_{\{i\}}) = \bfL_{ii}
\end{equation*}
 On the other hand, the unnormalized likelihood of selecting two items $\{i, j\}$ is moderated by the off-diagonal term $\bfL_{ij}$.
\begin{align*}
    \Pr[\text{Choose }\{i, j\}] &\propto \det\left(
    \begin{bmatrix}
    \bfL_{ii} & \bfL_{ij} \\
    \bfL{ji} & \bfL_{jj}
    \end{bmatrix}
    \right)\\
    &= \bfL_{ii} \bfL_{jj} - \bfL^2_{ij}
\end{align*}
The non-positive term $-\bfL^2_{ij}$ can be viewed as a penalty for subsets containing similar items, where the off-diagonal terms $\bfL_{ij}$ capture a notion of similarity between $i$ and $j$.

Decomposing the DPP kernel into similarity and quality components adds clarity to the model. \citet{KTstrucutedDPP} show that any DPP kernel $\bfL$ can be decomposed s.t. $\bfL_{ij}  = q_i \bfS_{ij}q_j$, where  $q_i$ is the \textit{quality} of item $i$, and $\bfS_{ij}$ the\textit{similarity} between items $i$ and $j$. The quality is viewed as the relative attractiveness of item $i$; the similarity quantifies the degree of similarity between the two items. Similarity terms are collected in a matrix, $\bfS$, satisfying $\bm{0} \preceq \bfS \preceq \bfI$\footnote{Here $\bfK \preceq \bfK'$ denotes that $\bfK' - \bfK $ is p.s.d.}. Under this decomposition, the DPP likelihood factors:
\begin{equation}
\label{eq:decomposed_DPP_likelihood}
\Pr[\text{Choose }C] \propto \left(\prod_{i \in C}q^2_i \right) \det(\bfS_C).
\end{equation}
The above formulation shows that the likelihood of a subset $C$ containing item $i$ is proportional to the squared quality of item $i$. A bias for diversity stems from the determinant of the similarity submatrix; the determinant takes values between $0$ and $1$. We leverage this decomposition to bridge the gap between DPPs and random utility subset choice models.

\section{Determinantal Processes \& Random Utility}
\label{sec:DPPs_RUM}
This section connects determinantal processes and random-utility-maximization-based subset choice models. In particular, we show that \emph{all} discrete determinantal point processes can be viewed as random-utility maximizing choices under a particular family of utility functions. Furthermore, studying this family of utility functions yields novel insights on the behavioral assumptions underlying DPPs, as well as connections between DPPs and existing subset choice models. This section takes a step back from the feature-based parametric view and considers discrete DPPs in their full generality; we return to features and parameters in \cref{sec:determinantal_choice}.

Our first result shows that samples from a DPP are random utility maximizing choices. The result is an affirmative answer to the questions ``is there a random utility function over subsets such that the utility-maximizing choice has the same distribution as a DPP''? The answer may seem obvious; the main idea is to ask the question.
\begin{theorem}
\label{thm:DPPSubsetChoice}
Let $A$ be a set of $n$ items and $\bfL$ an $n$-by-$n$ p.s.d. symmetric matrix. The following distributions over $2^A$ are equivalent:
\begin{enumerate}
    \item A DPP over $A$ with kernel $\bfL$.
    \item The utility-maximizing subset where each subset $C$ of $A$ has a log-determinant random utility
    \begin{equation*}
    U(C) = \log\det(\bfL_C) + \epsilon(C)
\end{equation*}
with $\epsilon(C)$ iid standard Gumbel errors.
\end{enumerate}
\end{theorem}
The proof of this and all subsequent results are deferred to  \cref{apx:proofs}. This theorem shows that a DPP is equivalent MNL over subsets, with a log-determinant utility.  We call this the \emph{utility implied by a DPP}.

 Simplification of the implied utility helps connect DPPs to existing subset choice models. Let $V(C)$ denote the non-random component of the implied utility of subset $C$. For a singleton, $C = \{i\}$, the non-random utility is equal to the log of the squared quality $v(i) := V(\{i\}) = \log(q^2_i)$. Decomposing the kernel matrix into quality and similarity terms as in \cref{eq:decomposed_DPP_likelihood} yields the following theorem.
\begin{theorem}
\label{thm:DPPAdditiveUtility}
    Let $\bfL$ be an $n$-by-$n$ DPP kernel. Then the non-random part of the induced utility of subset $C \subseteq A$ is
    \begin{equation*}
        V(C) = \sum_{i \in C}v(i) + \log\det(\bfS_C)
    \end{equation*}
\end{theorem}
In other words, one can view the implied utility as ``additive with corrections''. The correction in this case is a similarity-dependent penalty term $\log\det(\bfS_C)$. This term is non-positive because $\bfS \preceq \bfI$. Moreover, the DPP log-likelihood is submodular (\citet{KuleszaTaskarBook}, pp. 22).

Viewing a DPP as a subset choice model helps to connect DPPs to discrete choice and utility maximization. The fact that a DPP is  an MNL model over subsets immediately implies that DPPs obey the independence of irrelevant alternatives (IIA) property on a subset-level \citep{Train2009}; the relative probability of choosing subset $C$ over $C'$ does not depend on the presence of some third subset $C''$ in the assortment. This is a fundamental axiom of underlying many choice models. The utility function is also informative. Submodularity of the implied utility means that DPP choice exhibits diminishing marginal returns; the additional utility of adding an item to a selection is non-increasing in the size of the selection. Decreasing marginal utilities have long history in utility theory.

Our second result shows that the utility implied by a DPP is ``additive with corrections''. This helps highlight in what way DPPs differ from existing subset choice models. Indeed, DPPs enjoy novel properties from a subset choice perspective. To contrast with, e.g., the pairwise interaction model, the utility implied by a DPP is able to capture global dependencies, unrestricted by the size of the sets. Similarly, unlike the model of \citet{BensonSubset2018}, the DPP model does not restrict the number of sets that receive a correction. Instead, a DPP restricts the sign of the interactions it can capture; the corrections are all non-positive. This is a new approach in subset choice modeling, making DPPs a valuable addition to the subset choice modeling repertoire.

\section{A Determinantal Choice Model}
\label{sec:determinantal_choice}
This section defines a feature-based parametric DPP we call a \emph{determinantal choice model}. The main results of this section establish strong connections between this model, logistic regression, and MNL, in a way that lends the determinantal model components -- in particular the parameters -- intuitive and familiar interpretations. The approach is to show that the determinantal model family occupies a continuum of models, which subsumes logistic regression on the one extreme, and MNL on the other.

 The determinantal choice model is defined first. We take inspiration from conditional DPPs of \citet{Affandi2014Learning}, and determinantally thinned point processes of \citet{BlaszKeelerDetThinnig}. Instead of a fixed kernel, we specify a \emph{kernel function} that takes any assortment data $\bfX_A$ and generates a DPP kernel $\bfL(\bfX_A)$. We do this by specifying parametric quality and similarity models separately. Let $\bm\beta$ be a $d$-dimensional coefficient vector. The quality model is an exponentiated linear model \citep{KTstrucutedDPP}:
\begin{equation}
    q(\beta, \bfx_i) = \exp\left\{\frac{1}{2}\bm\beta^T \bfx_i \right\}.
\end{equation}
Under this quality model, the DPP likelihood is log-concave in the parameters $\bm\beta$ for any p.s.d. similarity model \cite{KTstrucutedDPP}. In this paper the similarity model is specified as an anisotropic RBF kernel with lengthscales $\bm\ell \in \mathbb{R}^d$:
 \begin{equation}
     S(\bm{\ell}, \bfx_i, \bfx_j) = \exp\left\{-\frac{1}{2}\sum^d_{k=1}\frac{|x_{i,k}- x_{j,k}|^2}{\ell^2_k}\right\}.
 \end{equation}
While we prefer the RBF for its interpretability, the results in this section are valid for any p.s.d. kernel function with $\bfS_{ij} \in [0, 1]$.  This completes the determinantal choice model;  the remainder of this section shows how the model relates to logistic regression and MNL, and discusses intuition for and interpretations of the parameters of this model.

The key insight of this section is to view subset choice as binary classification with dependent labels. The binary labels indicate selection; either an item is selected, $i \in C$, or it is not, $i \notin C$.  Thus, predicting a selected subset is equivalent to predicting binary labels for each item. The labels can be dependent. For example, in the case of negative dependence, conditioning on one item being selected may diminish the likelihood of other items being selected. Under this view, if the determinantal model has no similarity between distinct items, it should reduce to a conventional binary classification model that assumes independence between labels. Similarly, if all items are maximally negatively dependent -- that is mutually exclusive -- the determinantal model should resemble a classification model that selects at most one item.

Our first result formalizes this intuition in the case of totally dissimilar items. Under this similarity structure, the model is equivalent to a logistic regression.
\begin{theorem}
\label{thm:DPPisLogistic}
   Fix a finite assortment $A$ and data $\bfX_A$. If $\bfS(\bfX_A) = \bfI$, then the determinantal likelihood of $C \in A$ is
    \begin{equation*}
        \Pr[\text{Choose }C] = \prod_{i \in C}\left(\frac{e^{\bm\beta^T \bfx_i}}{1 + e^{\bm\beta^T \bfx_i}}\right)\prod_{j \in A \backslash C}\left(\frac{1}{1 + e^{\bm\beta^T \bfx_j}}\right).
    \end{equation*}
\end{theorem}
This is the likelihood of a logistic regression of \cref{eq:logit-likelihood}, treating the items in $A$ as independent observations. As a consequence of this result, one can interpret the coefficients $\bm\beta$ as logistic regression coefficients in this setting.

In the extreme of maximally negatively dependent labels, items are mutually exclusive. In this setting the determinantal model is equivalent to MNL over items.
\begin{theorem}
\label{thm:DPPisMNL}
    Fix a finite assortment $A$ and data $\bfX_A$. If $\bfS(\bfX_A) = \bfJ$ (the all $1$s matrix) then the determinantal likelihood of $\{i\}$ for $i \in C$ is
\begin{equation*}
        \Pr[\text{Choose }\{i\}] = \frac{\exp\{\bm\beta^T \bfx_i\}}{1 + \sum_{j \in C}\exp\{\bm\beta^T \bfx_j\}}
\end{equation*}
With $\Pr[\text{Choose } \emptyset] \propto 1$, and $0$ for all other sets.
\end{theorem}
The likelihood is precisely that of an MNL model over the assortment $A$ in \cref{eq:mnl-likelihood}. Analogously, whenever items are highly similar, the determinantal choice model parameters can be interpreted as MNL coefficients. Between the extremes, the interpretation still holds. One can always view the non-random utility, $\bm\beta^T\bfx_i$, the way one would under a logistic regression model by qualifying ``provided no other similar items be present''. When competing items are present, the marginal probability of choosing item $i$ is moderated by (a) the degree of negative dependence between $i$ and the competing items, and (b) the utility of the competing items. The exact probability, and marginal effects on it, can be evaluated by taking derivatives and evaluating assortments and other items at, e.g., mean values, as is common when interpreting MNL parameters.

\section{A comparison of models}
\label{sec:simulations}
The previous section shows that the determinantal choice model subsumes logistic regression and MNL whenever the similarity matrix takes on its extreme values. However, whether the model does this in practice is contingent on learning the ``correct'' parameters. This section uses synthetic data to examine how a trained determinantal choice compares with logistic regression and MNL as the degree of negative dependence in the training data is varied. Our experiment shows that the model behaves consistently with the theory; it indeed closely matches the performance of the classification models at the two extremes of negative dependence. Moreover, we find that determinantal choice model outperforms the two reference models when the degree of negative dependence lies between the two extremes.

\textbf{Experimental design.} The experiment estimates each model's out-of-sample predictive performance as a function of the degree of negative dependence in the selected subsets. This requires generating  a sequence of datasets with varying degrees of negative dependence. This is accomplished using a data generating process in which the negative dependence is parametrized by a scalar, $r > 0$. Each of the three models is parametrized over the same features, and trained over the same sequence of datasets. Evaluation data is withheld from each dataset during training. Finally, each model is evaluated on the same sequence of evaluation datasets.

\begin{figure}[ht]
\vskip 0.2in
\begin{center}
\centerline{\includegraphics[width=.6\columnwidth]{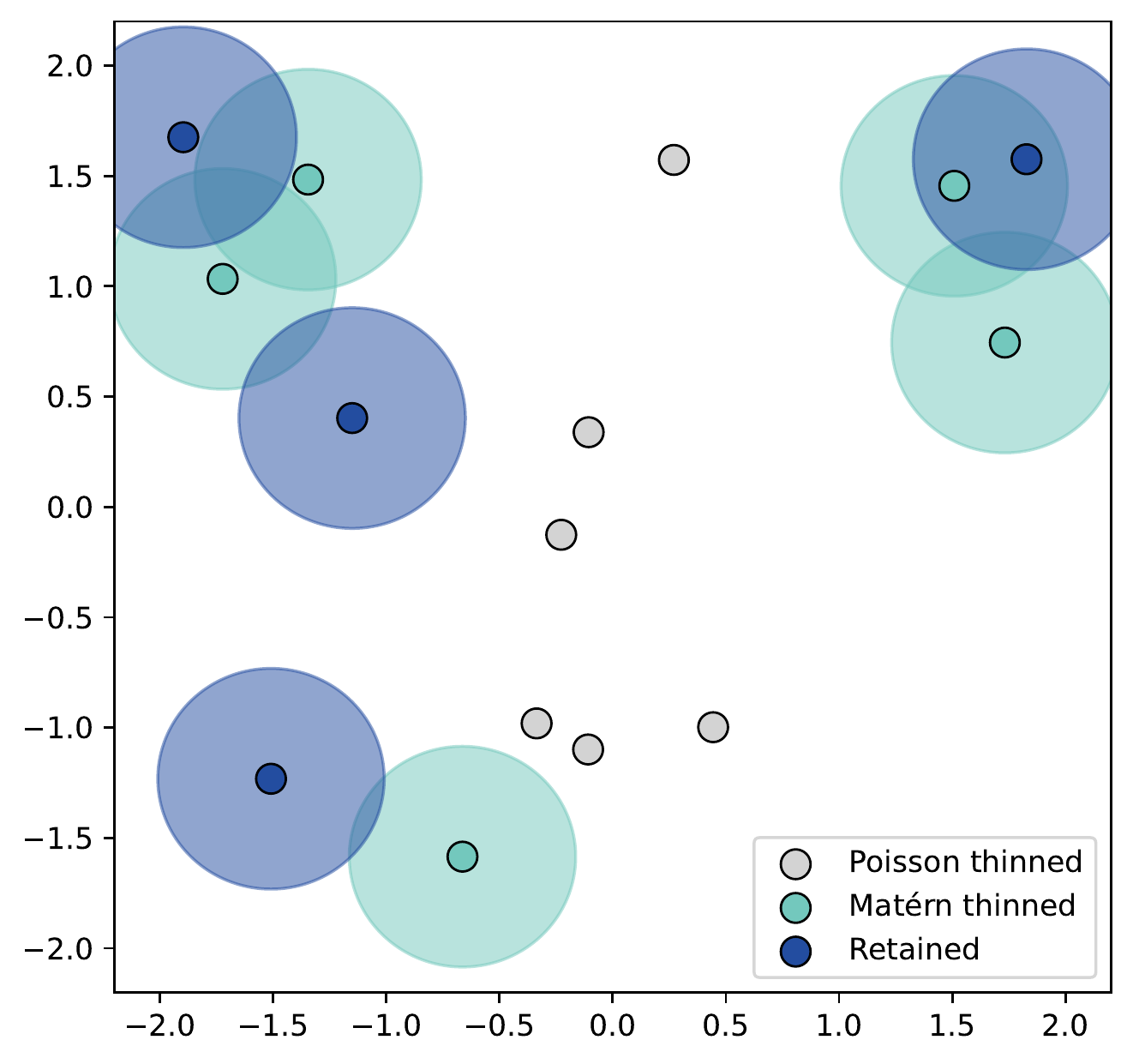}}
\caption{Illustration of one observation with 15 items and radius $r = 0.5$. Points in light grey are independently assigned $0$-labels in the first thinning phase. Light blue points are discarded during Mat{\'e}rn thinning; dark blue points retain their $1$-labels.}
\label{fig:simu_dgp}
\end{center}
\vskip -0.2in
\end{figure}

\textbf{Data generation.} The synthetic data is generated in three steps. Each observation -- an assortment-selection pair -- is generated independently. We describe the process for only one observation. First, the items and their features are sampled. Items are encoded as points in a $[-2, 2]^2$ square and are sampled uniformly at random. Each assortment has 15 items. Features $\bfX_i$ of the $i$th point are its $x$ and $y$ coordinates and the distance $d_i$ from the point to the origin.
Associated with each trial is a binary label vector $\bfy$ indicating which items are in the selection; $\bfy_i = 1$ denotes that $i$ is selected. The labels are initially set to $1$ for each item. The second sampling step sets the label of item $i$ to zero independently at random with probability
\begin{equation*}
    \Pr[\bfy_i = 0] =  1 - \min\left\{1, \exp\{\gamma_0 + \gamma_1 d_i\}\right\},
\end{equation*}
where $(\gamma_0, \gamma_1) = (-5, 2.5)$. The third and final step applies Mat{\'e}rn (\citeyear{Matern1960}) Type III thinning to the remaining 1-labeled points. This step takes as input a radius, $r > 0$. This is the source of negative dependence. Given a radius, $r > 0$, the remaining items with $1$-labels are sorted by their $y$-coordinates in decreasing order. The $1$-labeled item with the largest $y$-coordinate retains its $1$-label. Any neighboring items that are within $2r$ of this item have their labels set to $0$. The next largest item that still has a $1$ retains its $1$-label and its neighbors are given $0$-labels, and the process repeats until all items are processed. An example outcome is illustrated in \cref{fig:simu_dgp}. When the radius $r$ is small, labels are nearly independent; when it is large, the items become mutually exclusive. Repeating the process over a range of radii yields a sequence of datasets with varying negative dependence.

\textbf{Model and learning.} Learning takes a standard  Bayesian approach. An independent Gaussian prior is assumed over the quality coefficients $\bm\beta$. The same prior is used for the all three models. The similarity kernel function of the determinantal choice model takes only the distance between points as input, and its lengthscale $\ell$ is equipped with a log-normal prior. The log-likelihood is derived from  \cref{eq:decomposed_DPP_likelihood} and implemented in TensorFlow \citep{tensorflow2015-whitepaper}. Posterior samples are generated using MCMC with an adaptive kernel \citep{AndrieuAdaptiveMCMC, tfDistributions}. For each model-radius pair a total of 25 parallel MCMC chains are run. Predictions $\hat{\bfy}$ are formed by sampling parameters from the posterior distribution, and conditioning on these, sampling predictions from the DPP model using the DPPy implementation by \citet{DPPyTools}. The quality of a sampled prediction $\hat{\bfy}$ relative to true label vector $\bfy$ is quantified using the Matthews correlation coefficient $\mcc(\bfy, \hat{\bfy})$; this essentially measures correlation between binary vectors, and is known to be robust against label imbalance \citep{chicco20matthews}.

\begin{figure}[ht]
\vskip 0.2in
\begin{center}
\centerline{\includegraphics[width=.7\columnwidth]{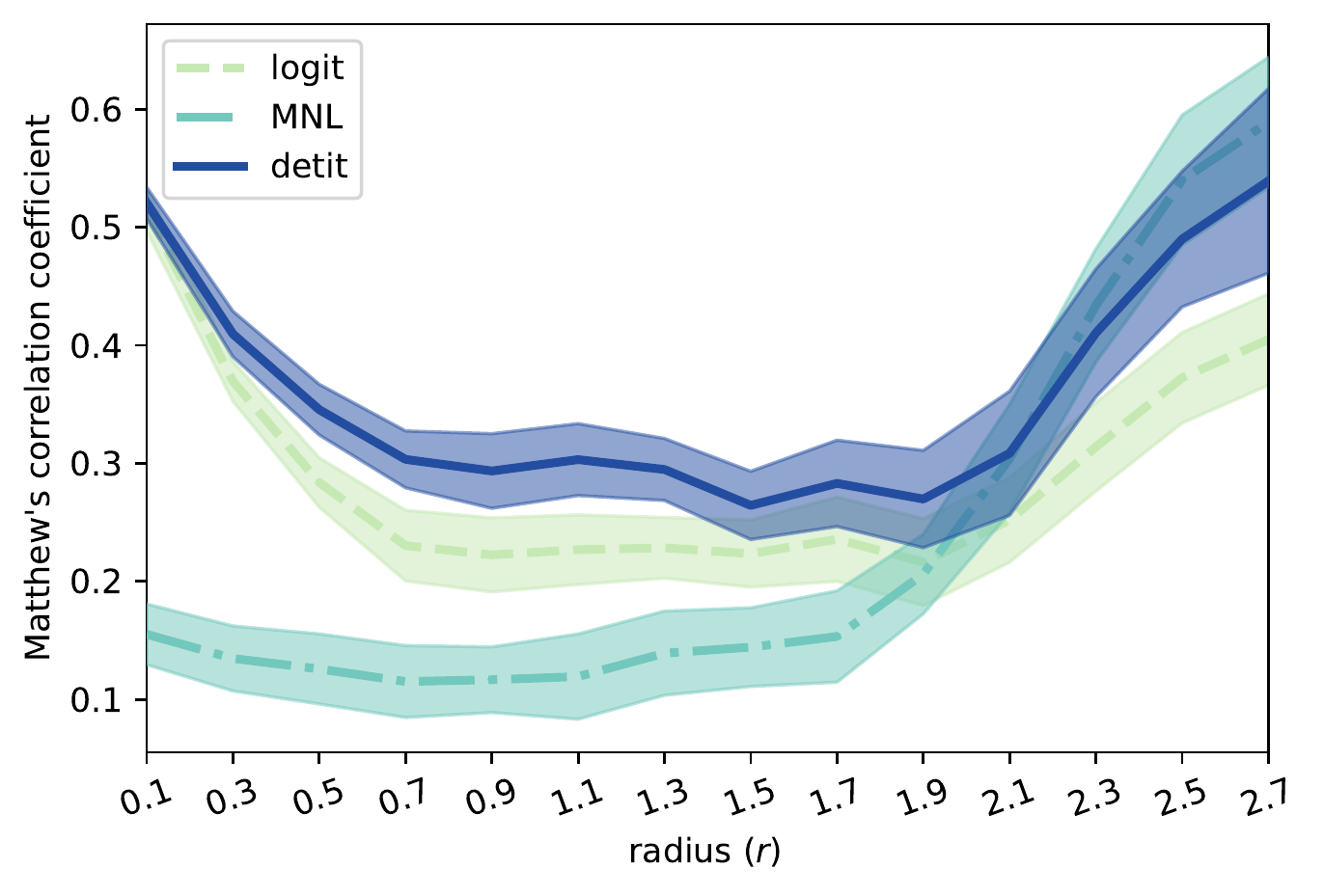}}
\caption{Mean Matthews correlation coefficients over varying radii $r$ for the logistic (light green -- dashed), MNL (dark green -- dotted), and determinantal choice model (dark blue -- solid), respectively. The shaded areas indicate $95\%$ confidence intervals.}
\label{fig:simu_matthews}
\end{center}
\vskip -0.2in
\end{figure}

\textbf{Results.} \cref{fig:simu_matthews} shows the Matthews correlation coefficient between predicted labels and true outcomes for each model, as a function of the radius. The results are averaged over posterior samples and assortments; the shaded regions denote 95\% CIs. On the left, the radius is small and item selections are nearly independent, the logistic and determinantal model predictions are strongly correlated with the true labels. The MNL predictions are less aligned with the true labels. This is due to the model predicting selections of size at most $1$ while the true selections contain multiple items on average. On the right, where the radii are large and items are nearly mutually exclusive, the MNL predictions attain a relatively high MCC. So does the determinantal model. At this extreme the logistic regression predictions are less correlated with the true labels since the model often predicts selections of multiple items while the true positive labels are nearly mutually exclusive. Note that throughout the range of experiments, the determinantal model's average performance is no worse than that of the two reference models.

\textbf{Discussion}. Our experiment shows that the determinantal choice model, when learned from data, can capture a range of negative dependence in the chosen subsets. This is in agreement with our theoretical results; the model behavior is consistent with logistic regression when there is little to no dependence in the data, and is consistent with MNL when dependence is maximally negative. Moreover, between the two extremes, when there is moderate negative dependence in subset selections, the model produces no worse predictions than the better of the two reference models. This suggests that the determinantal choice model indeed can learn a continuum of negative dependence structures.

\section{Wireless interference as subset choice}
\label{sec:application}
\textbf{Application.} In this section the determinantal choice model is employed to learn and validate a model of wireless interference in LoRaWANs using original experimental data. LoRaWAN or \emph{Long-Range Wide-Area Network} is a wireless communication protocol used in the \emph{Internet of Things}. There are 185 registered LoRaWAN operators and more than 225 million devices globally  \citep{LoRaAlliance, LoRaAllianceNDevs}. The main use-case for LoRaWAN is to enable small battery-driven end-devices, such as air-quality senors, to wirelessly transmit small packets of data to a receiver over distances up to a few kilometers \citep{LoRa-technical}. The communication protocol is spartan; it is designed to maximize the battery life of the end-devices. Consequently, devices do not coordinate transmissions, which creates potential for packet collisions -- the situation when two or more packets arrive at a receiver concurrently using the same radio parameters. Collisions may result in lost data. This can limit the scalability of LoRaWANs.  However, by understanding the severity of packet collisions, network planners can improve scalability via informed receiver placement and dimensioning. Understanding the mechanism of interference is key.

\textbf{Related work.} Many studies on LoRaWAN interference and scalability have been conducted \cite{Bor2016, Adelentado2017, GeorgiouRaza2017, Sundaram2019}. However, experimenting with physical networks at scale is infeasible; doing so would require thousands of end-devices and  would disrupt regular LoRaWAN users. Instead, the research focus is on building models for loss due to interference based on small-scale experiments and simulation. This has proven to be challenging. Most studies use only one or two devices -- if any -- to generate data. Furthermore, the models employed tend to be limited in their ability to capture interactions between colliding transmissions. Finally, few attempts are made to formally validate model performance on out-of-sample data.

This study treats LoRaWAN packet reception as subset choice, and learns an interference model from original experimental data. Our key insight is to view a set of incoming transmissions at a receiver in a given time-window as an assortment of items. Some subset of these transmissions are received (``chosen'') and others are lost. The transmissions are expected to exhibit negative interactions; if two transmission overlap in time and use the same radio parameters, is unlikely that all are successfully received. This is a consequence of limited receiver capacity; the receiver is said to ``lock on'' to the first incoming transmission, and to treat subsequent overlapping transmissions as noise \citep{LoRa-technical}.

\begin{figure}[ht]
\vskip 0.2in
\begin{center}
\centerline{\includegraphics[width=.7\columnwidth]{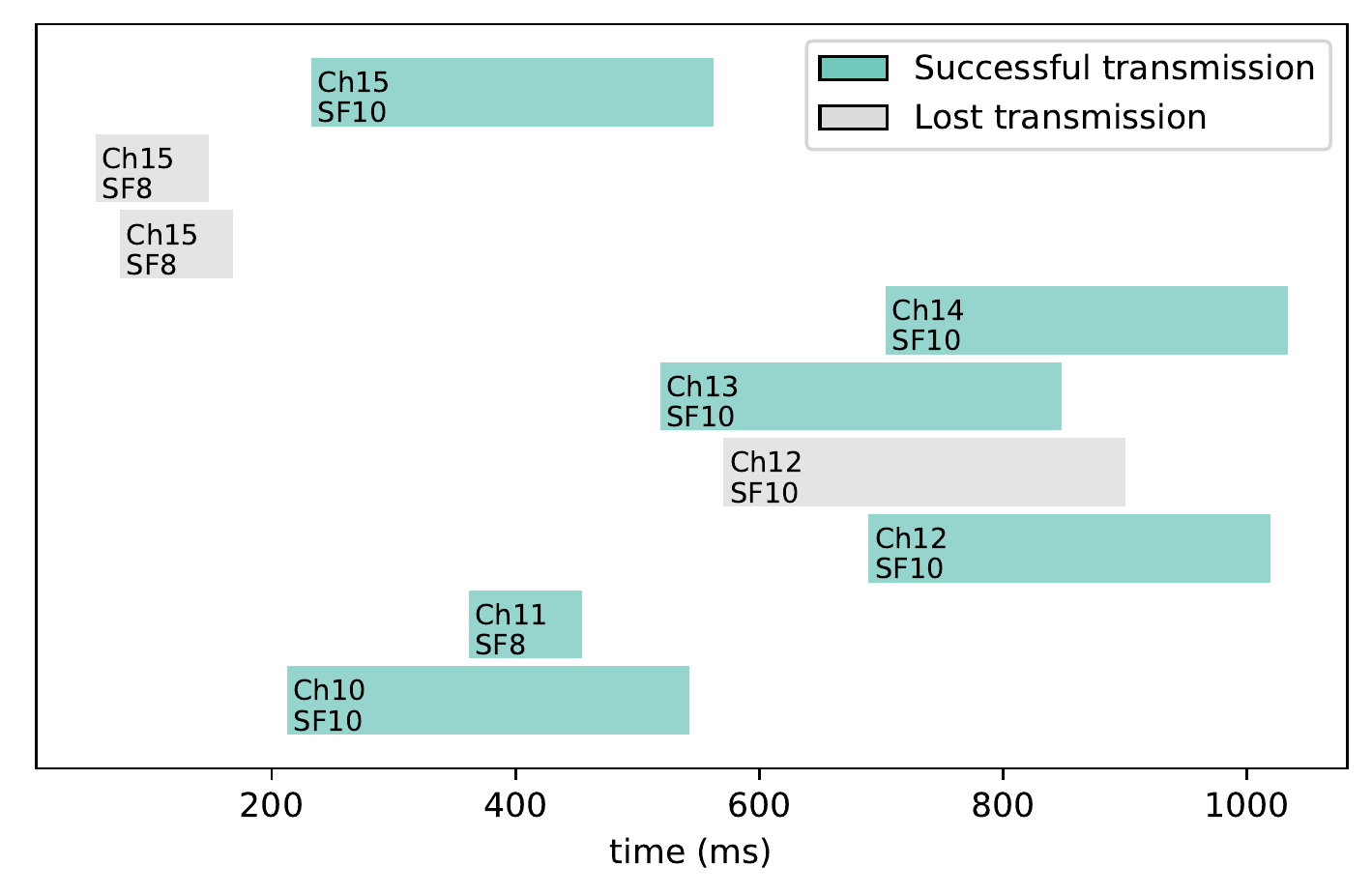}}
\caption{An example LoRaWAN trial. The bars represent transmissions. Their width and $x$-axis position indicates time-on-air. \texttt{Ch} stands for the channel, and \texttt{SF} for the spreading factor. Green transmission are received; grey ones are lost.} 
\label{fig:lora_trial}
\end{center}
\vskip -0.2in
\end{figure}

\textbf{Experiments and data.} Original data is generated using the following physical experiment. Up to nine end-devices transmit packets at randomly selected times, with randomly selected radio parameters. Each transmission is treated as an item, and its radio parameters as the features. Transmissions are grouped into time-disjoint sets; these are independent assortments. A LoRaWAN receiver is monitored to record the subset of transmissions, $C \subset A$, that successfully received for each assortment. An example assortment is shown in \cref{fig:lora_trial}. This setup is used to generate labeled subset choice data. \cref{apx:lora_experiment} contains more detailed descriptions of the devices and the experimental setup. A total of $N  = 1,030$ assortment-selection pairs are generated, of which $145$, all with at least one collision, are retained for evaluation.

There are a number of interesting features associated with each transmission. These are briefly described based on Semtech's (\citeyear{LoRa-technical}) technical summary. First, the \textsc{Channel} is a categorical variable taking integer values from $9$ to $16$. Transmissions on different channels are said to be orthogonal; a receiver should be able to receive transmissions on different channels concurrently, and transmissions on different channels should not interfere. The \textsc{Spreading factor} is another categorical variable ranging from $8$ to $11$. This determines the speed of a transmission; high spreading factors make signals more robust to noise at the cost of a longer time to transmit. Spreading factors are also considered orthogonal, even when on the same channel. Next, the \textsc{Power} (dBm) defines the strength of a transmission. The \textsc{Delay} (ms) describes the arrival time of a transmission, relative to an origin at $0$. Finally, the \textsc{Airtime} (ms) describes how long a transmission is on-air. Note that the spreading factor (SF) is the main determinant of the airtime in our setting; packets with similar SFs have similar airtimes.

\textbf{Model.} We specify a determinantal choice model using the above features and some closely associated variables. The quality model is given standardized versions of \textsc{Power} and \textsc{Delay}, as well as two dummy variables. The first dummy indicates the presence of a same-channel concurrent transmission; the second indicates the presence of a concurrent same-channel-same-SF transmission. Note that the latter implies the former. The similarity model takes a collection of dummy variables indicating the \textsc{Channel}, and a derived collection of continuous variables called \textsc{Relative delay}. The relative delay is the delay measured in packet airtimes rather than ms. Because airtime depends on the spreading factor, there are 4 dimensions to relative delay. If an item is on SF1, the relative delays corresponding to SF 2, 3, and 4 are set to zero, and so on. A large constant is further added to the non-zero entry. Thus, packets of different SFs are treated as dissimilar. While this approach is somewhat \textit{ad hoc}, this allows for improved modeling of the effect of time-overlap on mutual exclusivity\footnote{The perhaps most natural measure of packet similarity due to time-overlap is $[t_{\text{start}}, t_{\text{end}}]$ interval-overlap. This however does not generally produce a p.s.d. kernel function. The RBF with relative delays can be viewed as a ``smoothened'' version of this approach.}.

\begin{table}[t]
\caption{Posterior means, standard deviations, and effective sample sizes (ESS) of parameters. Similarity parameters are on log-scale.}
\label{tab:params-interference}
\vskip 0.15in
\begin{center}
\begin{small}
\begin{sc}
\begin{tabular}{lr@{}lccr}
\toprule
Variable & Posterior \  & mean & ESS & Model \\
\midrule
Constant    & 1.14 &$\pm$0.06& 19$K$ & Qual \\
Ch-overlap & -0.97 &$\pm$0.05& 20$K$ & Qual\\
Ch-Sf-overlap    &-0.84 &$\pm$0.04& 19$K$ & Qual \\
Power    & 0.51 &$\pm$0.05& 19$K$ & Qual        \\
Delay     & 1.07 &$\pm$0.02& 18$K$& Qual\\
Relative delay      & -0.62 &$\pm$0.01& 19$K$& Simi \\
Channel      & -1.24 &$\pm$1.06 & 456 & Simi        \\
\bottomrule
\end{tabular}
\end{sc}
\end{small}
\end{center}
\vskip -0.1in
\end{table}

\begin{figure}[ht]
\centering
\subfigure[Posterior histogram]{%
\label{fig:posterior}%
\includegraphics[width=.49\linewidth]{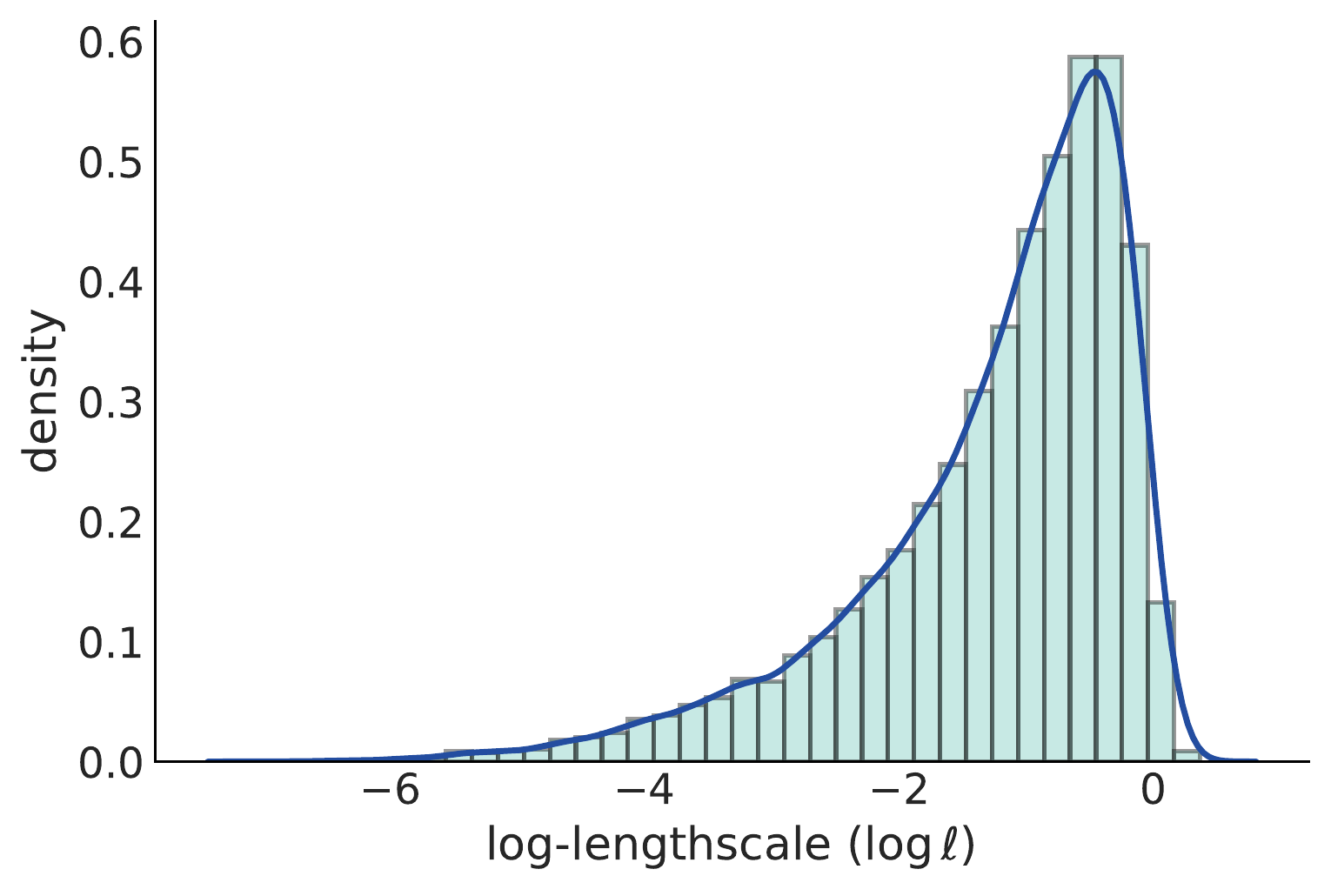}}%
\subfigure[Autocorrelation]{%
\label{fig:autocorr}%
\includegraphics[width=.49\linewidth]{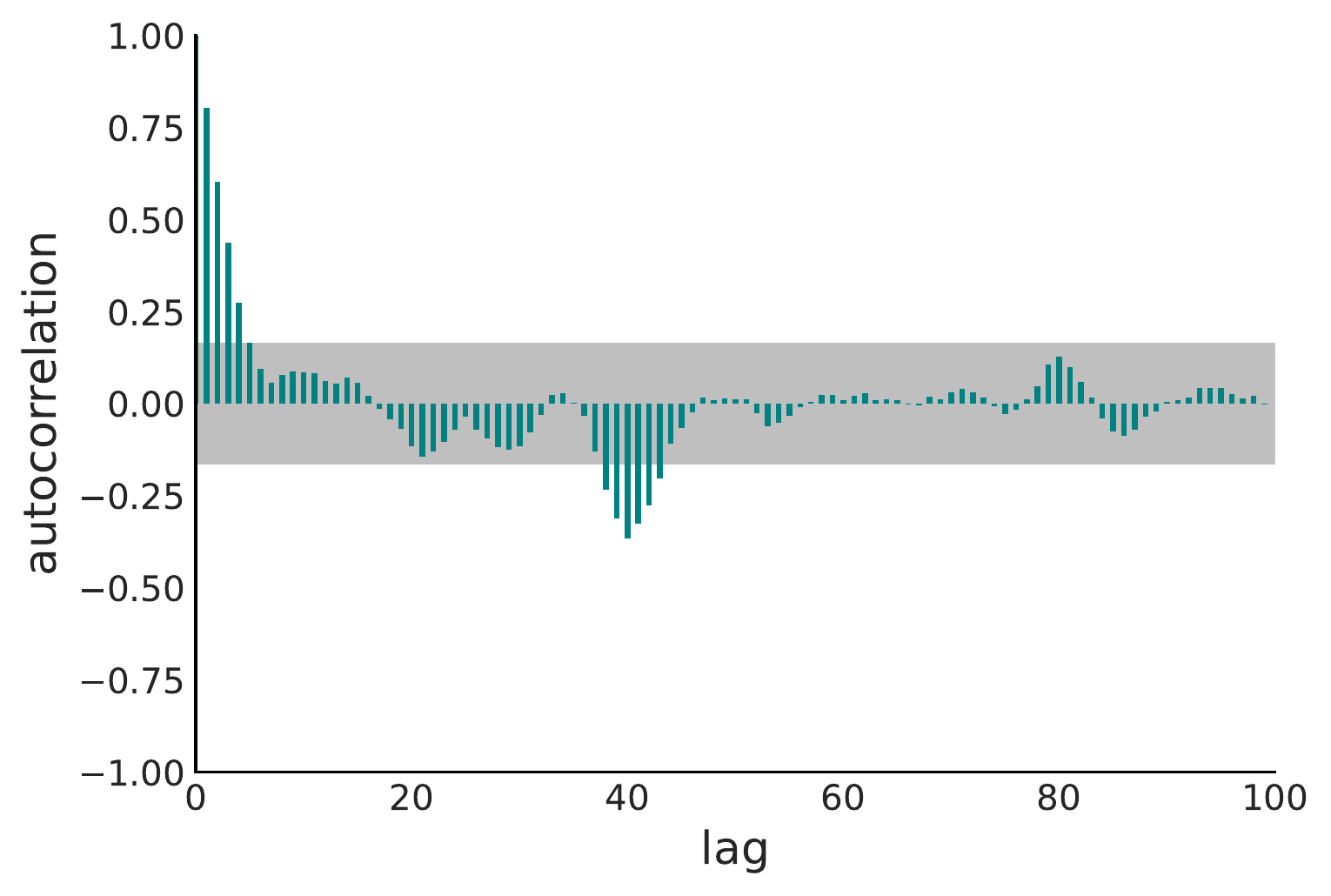}}%
\caption{Posterior histogram (a), and autocorrelation of a single MCMC chain (b) of the log-lengthscale of the \textsc{Channel} dummies. The latter plot is generated with ArviZ \citep{arviz_2019}.}
\end{figure}

\textbf{Bayesian inference.} Standard MCMC is used for Bayesian inference, as in \cref{sec:simulations}. Results are summarized in \cref{tab:params-interference}. The table columns contain, respectively, the variable name, posterior mean with standard errors, the effective sample size from MCMC, and the model component the variables belong to. The MCMC algorithm showed nominal convergence for all parameters, except the log-lengthscale of the channel dummies. The former parameters all show large effective sample sizes and  $\hat{R}$-values near $1$ \citep{Vehtari2021}. \cref{fig:posterior} summarizes the \textsc{Channel} log-lengthscale posterior, and \cref{fig:autocorr} the autocorrelation of one of its MCMC chains. The less than ideal convergence may be driven by autocorrelation over long lags. This means that the log-lengthscale samples may stem from a different distribution that the posterior that MCMC aims to approximate. This leaves room for potential improvement. Nevertheless, the other parameters are likely to stem distributions that approximate the true posterior well.

\textbf{Results.} Both the predictive performance and parameters are of interest. Overall, the Matthews correlation coefficient between sampled predictions and the true outcomes in the evaluation data -- averaged over the posterior -- is $0.25$ $\pm 0.03$. The posterior means of the parameters are summarized in  
\cref{tab:params-interference}. The parameters $\bm\beta$ of the quality model are all significantly different from $0$. The \textsc{Constant} is large and positive. Both coefficients of the \textsc{Power} and \textsc{Delay} are positive; the parameters of the dummy variables for concurrent same-channel overlaps are negative. The mean log-lengthscales of the similarity model are negative, although caution is to be used in interpreting the \textsc{Channel} lengthscale, due to the small number of effective samples.

\textbf{Interpretation.} Both the quality and similarity model parameters yield new insight into LoRaWAN interference. First, the coefficient of \textsc{Ch-Sf-overlap} is particularly interesting. The parameter is significantly negative. This indicates that the likelihood of a successful reception decreases if a same-channel concurrent transmission is present, \emph{despite being on a different spreading factor}.
This finding is in stark contrast with the claim that spreading factors are orthogonal, as this would imply a parameter value near zero. Secondly, the parameter associated with \textsc{Delay} is also surprising. It implies that arriving later improves the likelihood of success. However, under the view what a receiver ``locks on'' to the first signal it receives and treats subsequent signals as noise, arriving late should decrease the likelihood of a success. Finally, the log-lengthscales of the similarity models confirm our understanding of negative dependence in LoRaWAN. The lengthscale of the \textsc{Channel} seems reasonable; a negative log-lengthscale means that being on different channels renders packets \emph{dis}similar. This supports the view that channels are orthogonal, although the experiment is somewhat inconclusive with respect to this parameter. Finally, the negative log-lengthscale of the \textsc{Relative delay} implies that the similarity of two packets decays relatively fast as they become separated in time. At the posterior mean, two otherwise equal packets with $0.1$ airtime overlap have an similarity of roughly $0.25$. This implies the presence of a so-called \emph{capture effect} \citep{Rahmadhani-2018}; the the receiver seems able to receive two or more overlapping packets, provided they do not overlap too much. However, this finding may be partially driven by the imposed smoothness of the RBF kernel.

\textbf{Summary.} This study showcases the utility of our feature-based parameteric determinantal choice model on a learning task on original LoRaWAN interference data. The feature-based approach is essential as each observed assortment is unique. The model captures the observed negative dependence, and learns meaningful parameters. The parameters are indeed interpretable; contrary to a previous understanding of LoRaWAN, we find spreading factors to be non-orthogonal, and that arriving later to a collision is associated with a higher, rather than lower, likelihood of reception.

\section{Conclusion}
\label{sec:conclusion}
This paper shows new connections between determinantal point processes, and subset and discrete choice models. First, we show that all discrete DPPs are subset choice models with intuitive ``additive with corrections`` utility functions. In addition we specify a \emph{determinantal choice model} as a feature-based parametric determinantal process. This model is shown to capture a continuum of negative dependence between item labels. At the extreme of no dependence, we show that the model subsumes logistic regression; when dependence is maximal, the model reduces to MNL. Our simulation study shows that these connections hold even when the model is learned from data. Finally, we use our model to study wireless interference in LoRaWANs using original experiments, and draw new insights on LoRaWAN  interference by interpreting the model parameters. We believe the determinantal choice model can be of use in many interesting applications. 

\section*{Acknowledgments}
This material is based upon work supported by the National Science Foundation under Grant CNS-1952063.

\bibliographystyle{unsrtnat}
\bibliography{references}  






\newpage
\appendix
\section{Proofs of theorems}
\label{apx:proofs}
This section contains proofs for the theorems in the paper. 

\subsection{Proof of Theorem 3.1}
\textit{Proof of \cref{thm:DPPSubsetChoice}}. Fix a finite assortment $A$ and a p.s.d. symmetric kernel matrix $\bfL$ of size $|A|\times |A|$. Our goal is to prove that the following two distributions over $2^A$ are equivalent:
\begin{enumerate}
    \item A DPP over $A$ with kernel $\bfL$; and
    \item The random-utility maximizing choice where the utility of each subset $C \subseteq A$ is given by
    \begin{equation*}
        U(C) = \log\det(\bfL_C) + \epsilon(C),
    \end{equation*}
    with $\epsilon(C)$ iid standard Gumbel random errors. 
\end{enumerate}
It suffices to show that the likelihoods (\textit{pmf}s) are equivalent. First, recall that the likelihood of observing a subset $C \subseteq A$ under the DPP model is given by the DPP likelihood, 
\cref{eq:likelihood_kernel}:
\begin{equation*}
    \Pr_{\text{DPP}}[C] = \frac{\det(\bfL_C)}{\det(\bfI + \bfL)}.
\end{equation*}
For the random-utility maximizing choice we exploit the iid standard Gumbel errors to derive the softmax function, analogous to a standard random-utility derivation of MNL \citep{McFadden73}.
The probability that a fixed subset $C \subseteq A$ has the maximum utility is
\begin{equation*}
   \Pr_{\text{RUM}}[C] =  \Pr[U(C) > U(B) \text{ for all } B \subseteq A: B \neq C] = \frac{\exp\{\log\det(\bfL_C)\}}{\sum_{B \subseteq A}\exp\{\log\det(\bfL_C)\}}.
\end{equation*}
The above expression simplifies by cancelling the $\exp\{\log(\cdot)\}$ functions and recalling that the normalizing constant of a DPP satisfies the equality:
\begin{equation*}
    \sum_{B \subseteq A}\det(\bfL_B) = \det(\bfI + \bfL).
\end{equation*}
(See, e.g., Thm 2.1 in \citet{KuleszaTaskarBook}, pp. 8). These two observations yield the the final two equalities below:
\begin{equation*}
   \Pr_{\text{RUM}}[C]  = \frac{\exp\{\log\det(\bfL_C)\}}{\sum_{B \subseteq A}\exp\{\log\det(\bfL_C)\}} = \frac{\det(\bfL_C)}{\sum_{B \subseteq A}\det(\bfL_B)} = \frac{\det(\bfL_C)}{\det(\bfI + \bfL)}.
\end{equation*}
This proves that the two distributions over $2^A$ are equivalent. \hfill $\square$

\subsection{Proof of Theorem 3.2}
\textit{Proof of \cref{thm:DPPAdditiveUtility}}. Fix a finite assortment $A$ and corresponding DPP kernel $\bfL$. We need to prove that for any  subset of the assortment, $C \subseteq A$, the implied utility is equal to
\begin{equation*}
    V(C) = \log\det(\bfL_C) = \sum_{i \in C} v(i) + \log\det(\bfS_C),
\end{equation*}
where $\bfS$ is the similarity matrix satisfying $\bfL_{ij} = q_i \bfS_{ij} q_j$ for all $i, j \in A$, and $v(i) :=  \log\det(\bfL_{ii}) = 2\log(q_i)$.

The proof is nearly immediate after applying the DPP likelihood decomposition \cref{eq:decomposed_DPP_likelihood} to the proof above. First, the proof for \cref{thm:DPPSubsetChoice} establishes that the non-random component of the utility of a subset $C \subseteq A$ is $\log\det(\bfL_C)$. This is equal to the logarithm of the unnormalized DPP likelihood. Decomposing this likelihood inside the logarithm using \cref{eq:decomposed_DPP_likelihood} yields
\begin{equation*}
    V(C) = \log\det(\bfL_C) = \log\left(\left(\prod_{i \in C}q^2_i\right)\det(\bfS_C)\right) = \sum_{i \in C}2\log(q_i) + \log\det(\bfS_C).
\end{equation*}
Finally, recognizing that $v(i) \equiv 2\log(q_i)$ completes the proof. \hfill $\square$

\subsection{Proof of Theorem 4.1}
\textit{Proof of \cref{thm:DPPisLogistic}}. Fix an assortment $A$ with data $\bfX_A$, and a parameter vector $\bm\beta$. We need to prove that if $\bfS = \bfI$, then the determinantal choice model is equivalent to a logistic regression model on $\bfX_A$, in which the binary labels indicate an item $i$'s inclusion in the selection $C \subseteq A$. We prove the equivalence of the two models by showing that their likelihood functions are equivalent at all values of $(C, \bfX_A, \bm\beta)$, where $C \subseteq A$.

The likelihood of selecting items $C \subseteq A$ under a logistic regression model with data $\bfX_A$ and parameters $\bm\beta$ is given by the product of the probability of selecting the items in $C$, and the product of not selecting the items in $A \backslash C$:
\begin{equation*}
    \ell(C, \bfX_A, \bm\beta) = \prod_{i \in C}\left(\frac{e^{\bm\beta^T \bfx_i}}{1 + e^{\bm\beta^T \bfx_i}}\right) \prod_{j \in A \backslash C}\left(\frac{1}{1 + e^{\bm\beta^T \bfx_j}}\right).
\end{equation*}
The products are a consequence of the assumed independence between selections. To finish the proof, it suffices to show that the likelihood of the determinantal choice model, when $\bfS= \bfI$, is equal to this likelihood.

To this end, we make use the assumption that $\bfS = \bfI$. Under this assumption both the DPP kernel $\bfL$ and the the matrix $\bfL + \bfI$ are diagonal. This considerably simplifies the DPP likelihood expression.
\begin{equation*}
    \Pr[\text{Choose } C] = \frac{\det(\bfL_C)}{\det(\bfL + \bfI)} = \frac{\prod_{i \in C}\left(q^2(\bfx_i, \bm\beta)\right)}{\prod_{j \in A}\left(q^2(\bfx_j, \bm\beta) + 1\right)} = \prod_{i \in C}\left(\frac{q^2(\bfx_i, \bm\beta)}{q^2(\bfx_i, \bm\beta) + 1}\right)\prod_{j \in A \backslash C}\left(\frac{1}{q^2(\bfx_j, \bm\beta) + 1}\right)
\end{equation*}
The proof now follows from our particular choice of quality-model. Under this model the squared quality of item $i$ is exactly $e^{\bm\beta^T\bfx_i}$.
\begin{equation*}
    q^2(\bfx_i, \bm\beta) = \left(\exp\left\{\frac{1}{2}\bm\beta^T \bfx_i\right\}\right)^2 = \exp\{\bm\beta^T \bfx_i\}
\end{equation*}
\hfill $\square$

\subsection{Proof of Theorem 4.2}
\textit{Proof of \cref{thm:DPPisMNL}}. Fix an assortment $A$ with data $\bfX_A$, and a parameter vector $\bm\beta$. We need to prove that if $\bfS = \bfJ$ (the all $1$s matrix), then the determinantal choice model is equivalent to a MNL model in which the only outcomes with positive probability are the choice of a single item, $i \in A$, or opting out without choosing and item. It again suffices to prove that the likelihoods are equal, for an arbitrary selection $C \subseteq A$.

The likelihood function under an MNL model with feature matrix $\bfX_A$ and parameters $\bm\beta$ is
\begin{equation*}
    \Pr_{\text{MNL}}[\text{Choose } C] = 
    \begin{cases}
    \dfrac{\exp{\bm\beta^T \bfx_i}}{1 + \sum_{j \in C}\exp\{\bm\beta^T \bfx_j\}} & \text{if } C = \{i\} \text{ with } i \in A,\\
    \dfrac{1}{1 + \sum_{j \in C}\exp\{\bm\beta^T \bfx_j\}} & \text{if } C = \emptyset, \\
    0 & \text{otherwise.}
    \end{cases}
\end{equation*}

Now, consider the determinantal choice model likelihood of a singleton set $\{i\}$ with $i \in C$. As before the (unnormalized) likelihood of a singleton is the squared quality of that item.
\begin{equation*}
    \Pr_{\text{DPP}}[\text{Choose } \{i\}] \propto q^2(\bfx_i, \bm\beta) = \exp\{\bm\beta^T \bfx_i\}
\end{equation*}
Similarly, the unnormalized likelihood of choosing no item $1$. Finally, note that the likelihood of any subset $B \subseteq A$ of size two or larger is $0$, because the similarity matrix $\bfS_B$ is all $1$s, and thus perfectly linearly dependent, so its determinant is zero. It follows that the normalizing constant when $\bfS = \bfJ$ is exactly the sum of the squared qualities plus $1$ for the empty set.
\begin{equation*}
    \sum_{B \subseteq A}\det(\bfL_B) = 1  +\sum_{i \in C}\exp\{\bm\beta^T \bfx_i\}  
\end{equation*}
This proves that the determinantal choice model likelihood, when $\bfS = \bfI$, is equivalent to that of MNL. \hfill $\square$

\section{LoRaWAN Experimental Setup}
\label{apx:lora_experiment}

This section describes the laboratory LoRaWAN test bench and the process used to generate the interference data used in \cref{sec:application}. The main goal of the experiments conducted is to generate training data from which our determinantal choice model can learn the mechanisms underlying LoRaWAN interference, and evaluation data for judging the success of our approach. The test bench itself is also validated by replicating know LoRaWAN interference phenomena. This section proceeds with describing the equipment used, the test bench software, the validation experiments, and finally the generation of training and evaluation data.

\subsection{Equipment Used}
The test bench includes nine devices: eight Adafruit Feather M0+ LoRa boards with HopeRF
RFM95 modules; and one Sparkfun LoRa Thing Plus which includes a Semtech SX1262
LoRa radio. Antennas are a mix of commercially-available 2dBi whip antennas and
untuned wire whip antennas. The 2dBi antennas are not attached to a large ground-plane. An open-loop UART controls bus allows for simple hookup and synchronization
of the LoRa nodes. A Raspberry Pi 3B+ handled the interface with the SX1301 LoRa
concentrator.

\subsection{Test Bench Software}
All LoRa nodes are programmed with Arduino tools. The RFM95 devices use the radio driver packaged with MCCI's version of LMIC, while the SX1262 uses the RadioLib driver. A Python 3 serial interface allowed for control and logging of an experiment's transmitted packets, and libloragw's C interface allowed for logging of packets received on the SX1301.

\subsection{Validation Experiments}
Initial experiments are conducted to verify general functionality of the test bench. They
characterized the SX1301's ability to receive eight 125kHz bandwidth LoRa packets over different channels and spreading factors, despite transmission periods overlapping. Power sweep experiments characterized the
magnitude transfer function between the RFM95 and SX1301. There is a nearly linear relationship between transmitted power and received power in the test bed; this makes the transmitted power an effective estimate for received power. While received power -- the power at which a packet arrives a the receiver -- is of main interest, it is difficult to infer the received power of lost transmissions, as this is only registered upon successful reception. Transmitted power is a reasonable proxy in this setting. Overall, the experiments verify that the behavior of the test bench is broadly consistent with the expected behavior of LoRaWAN transmissions.


\subsection{Data generation}
The training and evaluation data is generated by specifying, and sampling from, a distribution over collections of LoRaWAN wireless transmissions. More specifically, a distribution over transmission parameters and transmission times is specified, sampled from, and then transmitted through the test bench. The parameter distribution is additionally manipulated to generate richer variability in the data, particularly by encouraging more collisions.

The data generating process is the following. There are $K > 0$ active devices. Each active device makes one transmission per observation. The transmission start time is sampled independently uniformly over $[0, D_{max}]$, rounded to the nearest integer. The value $D_{max} > 0$ is a maximum delay in ms, set to $2000$ ms.  Independently of the start time, each transmissions is assigned an integer-valued transmission power uniformly at random between between $-4$ and $23$ dBm. The channel and spreading factor are also chosen uniformly at random from a given set of $8$ and $4$ categorical values, respectively. To increase the occurrence of packet collisions,  the parameters are varied across some trials. In particular, for some trials $D_{max}$ is set to $600$ ms, and the number of channels and spreading factors is restricted to $2$ or $4$. The subsets of channels and spreading factors used are varied uniformly over observations whenever they are smaller than the full set of possibilities. Finally, the number of devices is  $K$ is varied between $7$ and $9$. This method was used to generate over $1,000$ observations. Because the focus of this work is on wireless interference in the presence of collisions, some observations (though not all) in which no collisions take place are removed  \textit{ex post}.


\end{document}